\documentclass[11pt]{article}

\usepackage[margin=1in]{geometry}
\usepackage[numbers,square,sort&compress]{natbib}
\usepackage{times}
\usepackage{microtype}
\usepackage{amsmath,amssymb}
\usepackage{graphicx}
\usepackage{booktabs}
\usepackage{multirow}
\usepackage{hyperref}
\usepackage{url}
\usepackage{xcolor}
\usepackage{listings}
\usepackage{caption}
\usepackage{subcaption}
\usepackage{setspace}
\usepackage{appendix}
\usepackage{placeins}
\usepackage[T1]{fontenc}
\usepackage[utf8]{inputenc}

\hypersetup{
  colorlinks=true,
  linkcolor=blue!60!black,
  citecolor=blue!60!black,
  urlcolor=blue!60!black
}

\title{Structured Synthetic Reasoning Data for\\
Arithmetic Fine-Tuning of Small Language Models}

\author{
  Jake O'Grady \\
  School of Computer Science \\
  University of Galway \\
  \texttt{j.ogrady13@universityofgalway.ie}
  \and
  Effirul Ramlan \\
  School of Computer Science \\
  University of Galway \\
  \texttt{effirul.ramlan@universityofgalway.ie}
}

\date{}

\begin{document}

\maketitle

\begin{abstract}
Small language models are attractive for local deployment, but they often struggle
with multi-step arithmetic reasoning. We study whether structured synthetic
reasoning data can improve this behaviour under consumer-hardware constraints.
Starting from GSM8K, we generated a 21{,}250-example corpus of grade-school
arithmetic word-problem variants using GPT-5-mini, 
combining natural-language solution traces, light
Socratic-style cues, structural variation, and irrelevant distractor context. We
then fine-tuned Qwen3-0.6B and Qwen3-1.7B with LoRA on consumer hardware
(Apple M4, 16\,GB RAM). Exact-match accuracy on GSM8K improved from 36.5\% to
49.1\% for Qwen3-0.6B and from 53.5\% to 66.5\% for Qwen3-1.7B. For Qwen3-1.7B, 
transfer to related arithmetic benchmarks was stronger, reaching
98.9\% on MultiArith and 73.0\% on SVAMP, compared with 54.4\% and 45.3\% for the
base model. Qualitative analysis suggests that fine-tuned models produce shorter
reasoning traces, make fewer arithmetic and distractor-use errors, and benefit
more consistently from self-consistency sampling. These results show that
low-cost synthetic data design can materially improve arithmetic adaptation in
small language models. Because the intervention combines Socratic-style cues with
other data-design choices, we interpret the gains as evidence for structured
synthetic reasoning data rather than as a causal test of Socratic guidance alone.
\end{abstract}


\section{Introduction}
\label{sec-introduction}

Large language models (LLMs) can solve a wide range of reasoning tasks, including
multi-step arithmetic~\citep{brown2020gpt3,kojima2022zeroshot}, and their
performance can be improved through prompting strategies such as chain-of-thought
reasoning~\citep{wei2022cot}. Smaller language models are more practical for local
deployment on consumer hardware, but they remain less reliable on tasks that require
multi-step calculation, operation selection, and resistance to irrelevant context.
The practical gap is therefore clear. The models most suitable for low-cost
deployment are also those most likely to fail on the reasoning behaviours that make
language models useful.

A growing body of work addresses this gap through parameter-efficient fine-tuning
(PEFT)~\citep{han2024peft}, retrieval-augmented generation~\citep{lewis2020rag}, and
reinforcement learning from human or automated
feedback~\citep{ouyang2022rlhf,havrilla2024rl}. PEFT methods such as
LoRA~\citep{hu2021lora} and QLoRA~\citep{dettmers2023qlora} have substantially
reduced the computational cost of model adaptation, making fine-tuning feasible
within modest hardware budgets. In parallel, teacher-generated synthetic datasets,
following the Self-Instruct paradigm~\citep{wang2022selfinstruct}, have shown that
useful training signal can be produced without large-scale manual annotation.

The structure of the training data also matters. Reasoning-focused supervision has
shown that intermediate reasoning traces can improve multi-step problem
solving~\citep{wei2022cot,wang2022selfconsistency}, while small models can be
especially sensitive to surface formatting in fine-tuning
data~\citep{zhou2023lima}. Educational approaches such as Socratic questioning and
the Feynman technique emphasise active, question-driven reasoning rather than
passive exposure to worked examples~\citep{delic2016socratic,reyes2021feynman,
perry2019metacognition}. Socratic-style prompting has also been explored at
inference time for LLM reasoning~\citep{qi2023socratic}. Less is known, however,
about how lightweight Socratic-style cues behave when embedded directly into
synthetic fine-tuning data for small language models.

This paper studies structured synthetic reasoning data for arithmetic adaptation in
small language models under consumer-hardware constraints. Starting from 
GSM8K~\citep{cobbe2021gsm8k}, we construct a filtered synthetic dataset that 
combines natural-language solution traces, light Socratic-style cues, structural variation, 
and deliberately irrelevant context. We then fine-tune Qwen3-0.6B and Qwen3-1.7B using 
LoRA and evaluate their performance on GSM8K, MultiArith, and SVAMP. These data-design 
choices are introduced together, so the results should be interpreted as evidence for 
structured synthetic reasoning data rather than as an isolated causal test of Socratic 
guidance alone.

Our contributions are as follows.
\begin{enumerate}
  \item We present a low-cost pipeline for constructing filtered synthetic arithmetic
        reasoning data from GSM8K, combining natural-language solution traces,
        light Socratic-style cues, structural variation, distractor context, and
        multi-stage filtering.

  \item We show that LoRA fine-tuning on this dataset improves exact-match accuracy
        for Qwen3-0.6B and Qwen3-1.7B on GSM8K, with stronger transfer to the related
        arithmetic benchmarks MultiArith and SVAMP.

  \item We analyse practical design lessons from the fine-tuning process, including
        the effect of solution formatting, LoRA layer coverage, and qualitative
        changes in arithmetic errors, distractor use, and reasoning concision.
\end{enumerate}

\section{Related Work}
\label{sec:related}

Parameter-efficient fine-tuning has made model adaptation feasible under more
modest compute budgets. LoRA~\citep{hu2021lora} freezes the pre-trained weights and
learns low-rank adapter matrices within selected transformer layers, reducing the
number of trainable parameters required for fine-tuning. QLoRA~\citep{dettmers2023qlora}
extends this approach by quantising the base model to 4-bit precision, allowing
larger models to be adapted under limited memory. AdaLoRA~\citep{zhang2023adalora}
further shows that adapter capacity need not be distributed uniformly across a
model, with feed-forward layers often receiving higher rank allocations than
attention layers. These methods provide the technical basis for adapting small and
quantised models within consumer-hardware constraints.

Synthetic data provides a complementary route to low-cost model adaptation.
DistilBERT~\citep{sanh2019distilbert} showed that compact student models can inherit
useful behaviour from larger teachers through distillation. Self-Instruct
~\citep{wang2022selfinstruct} and Stanford Alpaca~\citep{taori2023alpaca} later
showed that teacher-generated instruction data can support instruction following at
scale. In mathematical reasoning, MetaMath~\citep{yu2023metamath} and
WizardMath~\citep{luo2023wizardmath} demonstrated that GSM8K-derived augmentation
can substantially improve performance on arithmetic and mathematical word-problem
benchmarks. Our work follows this synthetic-data direction, but focuses on low-cost
data generation and fine-tuning for small models under local hardware constraints.

Prior work also shows that the form of the reasoning signal matters. Chain-of-thought
prompting~\citep{wei2022cot} improves multi-step reasoning by exposing models to
intermediate reasoning traces, while self-consistency sampling~\citep{wang2022selfconsistency} 
aggregates multiple reasoning paths to improve
answer selection. LIMA~\citep{zhou2023lima} suggests that small, high-quality
datasets can have a disproportionate effect, but also highlights the sensitivity of
smaller models to surface formatting in fine-tuning data. This is especially
important for arithmetic fine-tuning, where a model may learn the visible structure
of a solution format rather than the underlying operation-selection behaviour.

Socratic-style reasoning provides one possible way to structure this training signal.
Socratic questioning has been explored as an inference-time strategy for improving
LLM reasoning by decomposing problems into guiding sub-questions~\citep{qi2023socratic}. 
Related theoretical work on Socratic learning with language
games~\citep{schaul2024socratic} frames structured dialogue as a route towards
recursive improvement, although mainly in larger-scale settings. These studies
motivate the use of question-driven reasoning cues, but they do not establish how
such cues behave when embedded directly into synthetic fine-tuning data for small
language models. We therefore treat Socratic-style guidance as one component of a
structured synthetic reasoning dataset, rather than as an isolated causal mechanism.

\section{Method}
\label{sec:method}

\subsection{Dataset Construction}
\label{subsec:data}

We use GSM8K~\citep{cobbe2021gsm8k} as the source for constructing synthetic
training data and for primary held-out evaluation. GSM8K contains 7{,}473 training
problems with worked solutions and a test set of 1{,}319 problems, making it a
standard benchmark for grade-school arithmetic reasoning.

Each GSM8K training problem is submitted to GPT-5-mini
(\texttt{gpt-5-mini-2025-08-07}) through the OpenAI batch API. The generation prompt
asks the model to produce three variants of each source problem. Each variant is
required to vary the wording, structure, numerical values, or difficulty of the
source problem, include at least one sentence of irrelevant context, contain one or
two lightweight Socratic-style guiding questions within the solution narrative, and
end with a numeric answer marked by the \texttt{\#\#\#\#} delimiter used in
GSM8K~\citep{cobbe2021gsm8k}. The synthetic solutions are written as natural-language
reasoning traces with inline calculations rather than as rigid step lists. This
choice was made after an earlier dataset design caused the fine-tuned model to
reproduce surface templates rather than solve the underlying arithmetic task, as
discussed in Section~\ref{subsec:lessons}. Across all data-generation runs, 
the total API cost was \$20.88 for 12.8M tokens.

Generated outputs are processed through a four-stage filtering pipeline. The first
stage requires valid \texttt{Question:} and \texttt{Solution:} fields and a correctly
formatted \texttt{\#\#\#\#~<number>} final answer. The second removes trivially short
or excessively long solutions. The third applies 5-gram Jaccard deduplication with a
threshold of 0.85 to reduce near-duplicates. The fourth removes rhetorical questions
that occur immediately before the \texttt{\#\#\#\#} delimiter, because early
experiments showed that the model could reproduce this pattern at inference time.
The filtered dataset is split 90/10 into training and validation sets using random
seed 42, yielding 19{,}125 training examples and 2{,}125 validation examples.

\subsection{LoRA Fine-Tuning}
\label{subsec:finetuning}

Fine-tuning is performed using MLX LM~\citep{apple2024mlxlm}, Apple's machine
learning framework for Apple Silicon. The unified memory architecture allows the
available 16\,GB of RAM to be used for model training without a fixed CPU--GPU memory
split.

We fine-tune two models from the Qwen3 family~\citep{yang2025qwen3}. Qwen3-0.6B is
used in bfloat16 precision as the primary small-model target, chosen for rapid
iteration on consumer hardware. Qwen3-1.7B is used in 4-bit quantisation to test
whether the same pipeline can be applied to a larger model within the same memory
envelope.

LoRA adapters are applied to all 28 transformer layers. In each adapted layer, we
target the query ($\mathbf{W}_Q$), key ($\mathbf{W}_K$), value ($\mathbf{W}_V$), and
output ($\mathbf{W}_O$) projections in the self-attention module, together with the
gate, up, and down projections in the feed-forward module. For a pre-trained weight
matrix $\mathbf{W}_0 \in \mathbb{R}^{d \times k}$, LoRA represents the adapted layer
as

\begin{equation}
  h = (\mathbf{W}_0 + \Delta\mathbf{W})\,x = \mathbf{W}_0 x + \mathbf{B}\mathbf{A}x ,
  \label{eq:lora}
\end{equation}

where $\mathbf{A} \in \mathbb{R}^{r \times k}$ and
$\mathbf{B} \in \mathbb{R}^{d \times r}$ are trainable low-rank factors and
$r \ll \min(d,k)$. The base matrix $\mathbf{W}_0$ remains frozen. The matrix
$\mathbf{A}$ is initialised from $\mathcal{N}(0,\sigma^2)$ and $\mathbf{B}$ is
initialised to zero, so that $\Delta\mathbf{W}=0$ at the start of training.

All runs use AdamW with $\beta_1=0.9$, $\beta_2=0.98$, $\epsilon=10^{-6}$, and weight
decay $0.01$~\citep{loshchilov2019adamw}. We use cosine learning-rate decay with
linear warm-up~\citep{loshchilov2017sgdr}, batch size 4, gradient accumulation over
32 steps, effective batch size 128, and a maximum sequence length of 2{,}048 tokens.
Full hyperparameter details for all six fine-tuning runs are provided in
Appendix~\ref{app:configs}.

\subsection{Implementation Details}
\label{subsec:impl}

Model outputs are free-form text, so each output must be parsed to extract a single
numeric prediction for exact-match comparison. We use a regular expression that
searches for the last occurrence of the \texttt{\#\#\#\#} delimiter followed by an
integer or decimal number. The last match is used because baseline models sometimes
produce an intermediate estimate before restating the final answer. Numeric
predictions are normalised before comparison, so values such as \texttt{8.0} and
\texttt{8} are treated as equivalent. Outputs without a valid final-answer pattern
are recorded as empty predictions and scored as incorrect.

Training is capped at 6{,}000 iterations for Qwen3-0.6B runs and 10{,}000 iterations
for the largest Qwen3-1.7B run. Validation loss is evaluated every 200 steps on the
held-out validation split. In practice, validation loss stabilised around 5{,}000
steps, indicating that the iteration budgets were not binding. For each run, the
checkpoint with the lowest validation loss is used for reported evaluations rather
than the final checkpoint.

Evaluation results are written to a CSV file after each question. Each record stores
the question index, sampled outputs, extracted predictions, and whether each
prediction matches the ground-truth answer. If an evaluation run is interrupted, the
pipeline resumes from the last completed question and avoids duplicate scoring.

\subsection{Evaluation Protocol}
\label{subsec:eval}

The primary evaluation uses the GSM8K test set under 4-shot prompting, following the
original benchmark setup~\citep{cobbe2021gsm8k}. We report exact-match accuracy,

\begin{equation}
  \text{Accuracy} = \frac{\#\{A_s = A_g\}}{N},
  \label{eq:accuracy}
\end{equation}

where $A_s$ is the model prediction, $A_g$ is the ground-truth numeric answer, and
$N$ as the number of test cases. The primary GSM8K results use one sampled output 
per question unless otherwise stated.

For comparison, we evaluate the base Qwen3-0.6B and Qwen3-1.7B models alongside
several open-source baselines under the same prompting and answer-extraction
conditions. These include LLaMA-3.2-1B, LLaMA-3.2-3B-Instruct~\citep{grattafiori2024llama3}, 
Mistral-7B-Instruct-v0.3 in 8-bit quantisation~\citep{jiang2023mistral}, Gemma-3-4B-it~\citep{gemma2025}, 
and Qwen3-4B~\citep{yang2025qwen3}.

To test transfer and robustness, the fine-tuned models are also evaluated on two
benchmarks not used for training. SVAMP~\citep{patel2021svamp} contains 1{,}000 word
problems designed to test robustness to linguistic variation, while
MultiArith~\citep{roy2016multiarith} contains 180 compositional multi-step arithmetic
problems. For these evaluations, we report results with 1, 2, and 4
self-consistency samples~\citep{wang2022selfconsistency}, using temperature 0.7,
top-$p$ 0.95, top-$k$ 20, and majority-vote answer selection.

\section{Results}
\label{sec:results}

\subsection{GSM8K Performance}

Table~\ref{tab:baselines} reports 4-shot, single-sample exact-match accuracy on the
GSM8K test set. The fine-tuned Qwen3-0.6B model improves from 36.5\% to 49.1\%,
while the fine-tuned Qwen3-1.7B model improves from 53.5\% to 66.5\%.

\begin{table}[h]
  \centering
  \caption{GSM8K exact-match accuracy under 4-shot, single-sample evaluation.}
  \label{tab:baselines}
  \begin{tabular}{lrr}
    \toprule
    \textbf{Model} & \textbf{EM (\%)} & \textbf{Correct / Total} \\
    \midrule
    LLaMA-3.2-1B                     & 26.1 & 344 / 1319 \\
    Qwen3-0.6B (base)                & 36.5 & 481 / 1319 \\
    Mistral-7B-Instruct (8-bit)      & 47.2 & 622 / 1319 \\
    Qwen3-1.7B (base)                & 53.5 & 706 / 1319 \\
    LLaMA-3.2-3B-Instruct            & 56.4 & 744 / 1319 \\
    Gemma-3-4B-it                    & 75.1 & 991 / 1319 \\
    Qwen3-4B                         & 78.6 & 1037 / 1319 \\
    \midrule
    Qwen3-0.6B (fine-tuned, ours)    & \textbf{49.1} & 647 / 1319 \\
    Qwen3-1.7B (fine-tuned, ours)    & \textbf{66.5} & 877 / 1319 \\
    \bottomrule
  \end{tabular}
\end{table}

The fine-tuned 0.6B model outperforms the 8-bit Mistral-7B-Instruct baseline under
this evaluation setting, despite being much smaller. The fine-tuned 1.7B model also
exceeds the LLaMA-3.2-3B-Instruct baseline. These comparisons should be interpreted
with care, since the baselines differ in pre-training, instruction tuning,
quantisation, and model family. They nevertheless show that structured synthetic 
reasoning data can substantially improve arithmetic performance for the tested small models.

\subsection{LoRA Fine-Tuning Results}

Table~\ref{tab:qwen06} summarises three fine-tuning runs for Qwen3-0.6B. Run 1 adapts
only the first 12 of 28 transformer layers and does not improve over the base model.
Runs 2 and 3 adapt all 28 layers and both reach 49.1\% EM. Increasing the LoRA rank
from 16 to 32 does not improve performance under these settings.

\begin{table}[h]
  \centering
  \caption{Qwen3-0.6B LoRA fine-tuning results on GSM8K.}
  \label{tab:qwen06}
  \begin{tabular}{ccccc}
    \toprule
    \textbf{Run} & \textbf{Layers} & \textbf{Rank} & \textbf{Learning rate} & \textbf{EM (\%)} \\
    \midrule
    1 & 12 / 28 & 16 & $1 \times 10^{-4}$ & 36.5 \\
    2 & 28 / 28 & 16 & $8 \times 10^{-5}$ & \textbf{49.1} \\
    3 & 28 / 28 & 32 & $5 \times 10^{-5}$ & \textbf{49.1} \\
    \bottomrule
  \end{tabular}
\end{table}

Table~\ref{tab:qwen17} reports the corresponding runs for Qwen3-1.7B. All three runs
adapt all 28 layers. The best result is obtained with rank 16 and learning rate
$1 \times 10^{-4}$, reaching 66.5\% EM. Higher ranks do not improve performance in
this setting, which suggests that adapter capacity, quantisation, and data size may
interact in non-trivial ways.

\begin{table}[h]
  \centering
  \caption{Qwen3-1.7B LoRA fine-tuning results on GSM8K.}
  \label{tab:qwen17}
  \begin{tabular}{ccccc}
    \toprule
    \textbf{Run} & \textbf{Layers} & \textbf{Rank} & \textbf{Learning rate} & \textbf{EM (\%)} \\
    \midrule
    1 & 28 / 28 & 48 & $4 \times 10^{-5}$ & 61.0 \\
    2 & 28 / 28 & 32 & $5 \times 10^{-5}$ & 60.7 \\
    3 & 28 / 28 & 16 & $1 \times 10^{-4}$ & \textbf{66.5} \\
    \bottomrule
  \end{tabular}
\end{table}

\subsection{Transfer to Related Arithmetic Benchmarks}

Tables~\ref{tab:multiarith} and~\ref{tab:svamp} report exact-match accuracy on
MultiArith and SVAMP for the base and fine-tuned Qwen models under 1, 2, and 4
self-consistency samples.

\begin{table}[h]
  \centering
  \caption{MultiArith exact-match accuracy across self-consistency sampling budgets.}
  \label{tab:multiarith}
  \begin{tabular}{llr}
    \toprule
    \textbf{Model} & \textbf{Samples} & \textbf{Accuracy (\%)} \\
    \midrule
    Base Qwen3-0.6B       & 1 & 45.0 \\
    Base Qwen3-0.6B       & 2 & 41.0 \\
    Base Qwen3-0.6B       & 4 & 60.0 \\
    \midrule
    Fine-tuned Qwen3-0.6B & 1 & 79.4 \\
    Fine-tuned Qwen3-0.6B & 2 & 79.4 \\
    Fine-tuned Qwen3-0.6B & 4 & \textbf{92.2} \\
    \midrule
    Base Qwen3-1.7B       & 1 & 48.3 \\
    Base Qwen3-1.7B       & 2 & 41.7 \\
    Base Qwen3-1.7B       & 4 & 54.4 \\
    \midrule
    Fine-tuned Qwen3-1.7B & 1 & 94.4 \\
    Fine-tuned Qwen3-1.7B & 2 & 97.2 \\
    Fine-tuned Qwen3-1.7B & 4 & \textbf{98.9} \\
    \bottomrule
  \end{tabular}
\end{table}

\begin{table}[h]
  \centering
  \caption{SVAMP exact-match accuracy across self-consistency sampling budgets.}
  \label{tab:svamp}
  \begin{tabular}{llr}
    \toprule
    \textbf{Model} & \textbf{Samples} & \textbf{Accuracy (\%)} \\
    \midrule
    Base Qwen3-0.6B       & 1 & 29.0 \\
    Base Qwen3-0.6B       & 2 & 33.0 \\
    Base Qwen3-0.6B       & 4 & 42.3 \\
    \midrule
    Fine-tuned Qwen3-0.6B & 1 & 45.3 \\
    Fine-tuned Qwen3-0.6B & 2 & 43.7 \\
    Fine-tuned Qwen3-0.6B & 4 & \textbf{52.7} \\
    \midrule
    Base Qwen3-1.7B       & 1 & 39.7 \\
    Base Qwen3-1.7B       & 2 & 33.3 \\
    Base Qwen3-1.7B       & 4 & 45.3 \\
    \midrule
    Fine-tuned Qwen3-1.7B & 1 & 65.3 \\
    Fine-tuned Qwen3-1.7B & 2 & 67.0 \\
    Fine-tuned Qwen3-1.7B & 4 & \textbf{73.0} \\
    \bottomrule
  \end{tabular}
\end{table}

The strongest transfer is observed on MultiArith. With four self-consistency samples,
Qwen3-0.6B improves from 60.0\% to 92.2\%, while Qwen3-1.7B improves from 54.4\% to
98.9\%. The fine-tuned models also benefit more consistently from additional
self-consistency samples than the base models. This is consistent with the fine-tuned
models producing a more reliable distribution of arithmetic outputs.

The gains on SVAMP are smaller but still substantial. With four samples, Qwen3-0.6B
improves from 42.3\% to 52.7\%, and Qwen3-1.7B improves from 45.3\% to 73.0\%.
The weaker transfer to SVAMP suggests that the GSM8K-derived training data improves
arithmetic problem solving more strongly than robustness to varied linguistic
formulations. They nevertheless suggest that the proposed pipeline can produce small 
models that are competitive with substantially larger baselines on GSM8K

\FloatBarrier

\section{Qualitative Analysis}
\label{sec:qualitative}

The following examples are illustrative rather than a systematic error analysis. They
show how the fine-tuned models differ from their base counterparts in arithmetic
consistency, distractor use, and output concision.

\subsection{MultiArith}

Consider the following problem.

\begin{quote}
  \textit{A company invited 47 people to a luncheon, but 7 of them didn't show up. If
  the tables they had held 5 people each, how many tables do they need?}
\end{quote}

The baseline Qwen3-0.6B exhibits several failure patterns across four sampled outputs.
These include an arithmetic error, where $47 - 7$ is computed as 33, a mid-generation
shift to an unrelated school cafeteria problem, and an incoherent final answer after
partly correct intermediate reasoning. Only one of the four samples is correct.

The fine-tuned model is more consistent on the same problem. All four sampled outputs
correctly compute $47 - 7 = 40$ and $40 \div 5 = 8$. The responses are also shorter,
with reasoning largely confined to the required calculation. Three of the four samples
produce the correct final answer, with the remaining error occurring during a
verification step rather than in the main calculation.

\subsection{SVAMP}

Consider a SVAMP problem containing irrelevant information.

\begin{quote}
  \textit{There are 41 bird families living near the mountain. If 35 bird families flew
  away to Asia and 62 bird families flew away to Africa, how many more bird families
  flew away to Africa than those that flew away to Asia?}
\end{quote}

The correct operation is $62 - 35 = 27$. The quantity 41 is irrelevant.

Both the base and fine-tuned models answer two of four sampled outputs correctly on
this example. Their error patterns differ, however. The baseline model repeatedly
incorporates the irrelevant quantity 41, performs unnecessary secondary operations, and
in two samples truncates before reaching a final answer. The fine-tuned model ignores
the distractor in all four outputs and performs only the required subtraction. This
example is consistent with the broader SVAMP results, where fine-tuning improves
performance but does not eliminate errors under linguistically varied problem
formulations.

\section{Dataset and Fine-Tuning Design Lessons}
\label{subsec:lessons}

The experiments also exposed practical design issues that affected small-model
fine-tuning. These observations are not separate benchmark results, but they are
important for interpreting why the final dataset and LoRA configuration were used.

\paragraph{Format sensitivity in small models.}
Our initial dataset used a rigid checklist-and-numbered-question format, decomposing
each solution into a conceptual checklist followed by numbered Socratic questions with
inline calculations. When fine-tuned on this data, Qwen3-0.6B learned to reproduce the
surface structure of the format rather than solve the underlying arithmetic task. On
the unstructured GSM8K test set, the model often attempted to impose the checklist
template onto problems that did not follow it. This behaviour is consistent with
prior evidence that small language models can be highly sensitive to surface
formatting in fine-tuning data~\citep{zhou2023lima}.

\paragraph{Natural-language reasoning traces.}
We therefore replaced the rigid format with natural-language solutions in which
Socratic-style cues appear as short guiding questions within the reasoning narrative.
This avoided a fixed external template and reduced format overfitting in our
subsequent runs. Since this change was made together with other dataset revisions, it
should be interpreted as a design lesson rather than as an isolated ablation.

\paragraph{Feed-forward layer coverage.}
Early runs applied LoRA adapters only to self-attention projections and omitted the
feed-forward sublayers. These runs performed substantially worse than runs that
adapted both self-attention and feed-forward projections. This observation is
consistent with AdaLoRA~\citep{zhang2023adalora}, which finds that feed-forward
modules often require substantial adapter capacity. In our final configuration,
adapters are applied to the query, key, value, output, gate, up, and down projections
across all adapted transformer blocks.

\section{Discussion}
\label{sec:discussion}

The main result is that structured synthetic reasoning data can substantially improve
arithmetic adaptation in small language models under consumer-hardware constraints.
Fine-tuning Qwen3-0.6B improves GSM8K accuracy by 12.6 percentage points, and the
same model reaches 92.2\% on MultiArith with four self-consistency samples. The
Qwen3-1.7B model shows larger transfer gains, reaching 98.9\% on MultiArith and
73.0\% on SVAMP. These results indicate that small models can acquire useful
arithmetic behaviour from a relatively low-cost synthetic data pipeline.

The results do not establish that Socratic-style guidance alone causes the observed
improvements. The intervention combines several data-design choices, including
natural-language reasoning traces, structural variation, irrelevant distractor
context, answer-format filtering, deduplication, and lightweight Socratic-style cues.
A clean ablation against an otherwise identical non-Socratic synthetic dataset would
be required to isolate the contribution of the guiding questions. The safer
interpretation is that Socratic-style cues are a plausible component of a broader
structured synthetic-data design.

Transfer is strongest on MultiArith and weaker on SVAMP. This pattern suggests that
the GSM8K-derived training data improves arithmetic operation selection and
multi-step calculation more strongly than robustness to varied linguistic
formulations. Improving SVAMP performance may require greater linguistic diversity in
the synthetic data, including more varied surface forms, distractor types, and problem
templates.

The LoRA experiments also show that adapter configuration matters. For Qwen3-0.6B,
adapting all layers is more important than increasing rank from 16 to 32. For the
4-bit Qwen3-1.7B model, the best result is obtained with the lowest rank tested. This
may reflect regularisation under quantisation, but the present experiments are not
sufficient to make a causal claim. More systematic rank, learning-rate, and
quantisation sweeps would be needed to characterise this interaction.

\paragraph{Limitations.}
This study has several limitations. First, the evaluation is restricted to arithmetic
word-problem benchmarks, so the results do not establish general improvements in
commonsense, symbolic, or logical reasoning. Second, the synthetic dataset is derived
from GSM8K, which may limit the strength of claims about out-of-domain
generalisation. Third, the current study does not include a matched non-Socratic
synthetic-data baseline, so the effect of Socratic-style cues cannot be isolated from
the broader data-generation and filtering process. Fourth, the dataset is generated
using a single teacher model, GPT-5-mini, and may reflect that model's style and
failure modes. Fifth, the experiments are conducted on one consumer-hardware setup,
so runtime, memory behaviour, and training stability may differ on other systems.
Finally, the qualitative examples are illustrative. A systematic error analysis would
be needed to quantify changes in arithmetic errors, distractor uptake, truncation, and
answer-format failures.

\section{Conclusion}
\label{sec:conclusion}

This paper shows that structured synthetic reasoning data can substantially improve
arithmetic adaptation in small language models under consumer-hardware constraints.
Fine-tuning Qwen3-0.6B on a 21{,}250-example GSM8K-derived dataset improves
exact-match accuracy on GSM8K by 12.6 percentage points and reaches 92.2\% on
MultiArith with four self-consistency samples. The fine-tuned Qwen3-1.7B model
achieves 66.5\% on GSM8K, 98.9\% on MultiArith, and 73.0\% on SVAMP. These results
show that low-cost synthetic data generation and LoRA fine-tuning can produce
meaningful arithmetic gains in locally runnable models.

The results should be interpreted as evidence for structured synthetic reasoning data
rather than as a causal test of Socratic-style guidance alone. The intervention
combines natural-language reasoning traces, lightweight Socratic-style cues,
structural variation, distractor context, filtering, and deduplication. Future work
should therefore include matched ablations against non-Socratic synthetic data,
broader evaluation beyond arithmetic word problems, and reinforcement learning from
correctness-based rewards after supervised fine-tuning. These extensions would clarify
which parts of the data design drive the observed gains and whether the approach
generalises beyond arithmetic reasoning.

\section*{Data and Code Availability}
Code, configuration files, data-generation prompts, and processed evaluation logs
are available at \url{https://github.com/jakeogrady/socratiq}. The
underlying GSM8K, MultiArith, and SVAMP benchmarks are publicly available from
their original sources.

\bibliographystyle{unsrtnat}
\bibliography{cas-refs}
\newpage

\appendix

\section{Data Generation Prompt}
\label{app:prompt}

\begin{verbatim}
You are a math tutor tasked with generating training data by rewriting
math solutions into multiple concise reasoning variants, using gentle
Socratic-style guidance.

Rules:
- Generate 3 altered questions similar in structure and difficulty,
  each using basic Socratic questioning to improve accuracy.
- Each question MUST contain at least one sentence of redundant
  information not used in the solution.
- Each problem MUST be unique, with a different structure.
- Use diverse language; vary phrasing and logical steps.
- Provide a brief, natural language reasoning solution with light
  calculations embedded.
- Each pair must begin with "Question:" and "Solution:".
- Do not use bullet points, numbered lists, or structured step trees.
- Include gentle Socratic curiosity, with 1-2 short guiding questions
  per solution.
- Do NOT ask rhetorical questions after presenting the final answer.
- Answers must always be positive integers.
- Each solution MUST always have a numeric answer formatted as:
  #### answer
- Insert <|endofsolution|> after each question-solution pair.
- All counts of physical objects must remain >= 0 at every step.
\end{verbatim}

\section{LoRA Configuration Summary}
\label{app:configs}

\begin{table}[htbp]
  \centering
  \caption{Summary of fine-tuning run configurations.}
  \label{tab:configs}
  \begin{tabular}{llccccc}
    \toprule
    \textbf{Model} & \textbf{Run} & \textbf{Layers} & \textbf{Rank} &
    \textbf{LoRA scale} & \textbf{Learning rate} & \textbf{Iterations} \\
    \midrule
    \multirow{3}{*}{Qwen3-0.6B}
      & 1 & 12 & 16 & 24.0 & $1 \times 10^{-4}$ & 6{,}000 \\
      & 2 & 28 & 16 & 24.0 & $8 \times 10^{-5}$ & 6{,}000 \\
      & 3 & 28 & 32 & 32.0 & $5 \times 10^{-5}$ & 6{,}000 \\
    \midrule
    \multirow{3}{*}{Qwen3-1.7B}
      & 1 & 28 & 48 & 40.0 & $4 \times 10^{-5}$ & 10{,}000 \\
      & 2 & 28 & 32 & 32.0 & $5 \times 10^{-5}$ & 6{,}000 \\
      & 3 & 28 & 16 & 8.0  & $1 \times 10^{-4}$ & 6{,}000 \\
    \bottomrule
  \end{tabular}
\end{table}

All runs use AdamW with $\beta_1=0.9$, $\beta_2=0.98$, $\epsilon=10^{-6}$, and
weight decay $0.01$. Training uses cosine learning-rate decay with a 500-step
warm-up, batch size 4, gradient accumulation over 32 steps, maximum sequence length
2{,}048, and gradient checkpointing. LoRA adapters are applied to
\texttt{q\_proj}, \texttt{k\_proj}, \texttt{v\_proj}, \texttt{o\_proj},
\texttt{gate\_proj}, \texttt{up\_proj}, and \texttt{down\_proj} in all adapted
transformer blocks.

\end{document}